\documentclass[10pt,twocolumn,letterpaper]{IEEEconf}

\usepackage{graphicx}
\usepackage{amsmath}
\usepackage{amssymb}
\usepackage{amsfonts}
\usepackage{bm}
\usepackage{booktabs}
\usepackage{makecell}
\usepackage{siunitx}
\usepackage{textcomp}
\usepackage{xcolor}
\usepackage{url}
\usepackage{cite}
\usepackage[colorlinks=true,citecolor=blue,linkcolor=red,urlcolor=blue]{hyperref}

\let\oldthebibliography\thebibliography
\renewcommand{\thebibliography}[1]{%
  \oldthebibliography{#1}%
  \setlength{\itemsep}{0pt}%
  \setlength{\parskip}{0pt}%
}

\title{ACEsplat: Accelerated 3D Gaussian Scene Regression via RGB \\ and Poses Only}

\author{Mingkai Liu$^{1,\dagger}$, Haohua Que$^{2,6,\dagger}$, Dikai Fan$^{3}$, Haojia Gao$^{4}$, Tianle Zhu$^{2}$, Handong Yao$^{2}$, Qian Zhang$^{4,6}$, \\
Ruopeng Zhang$^{5}$, Xianliang Huang$^{3}$, Fei Qiao$^{4,*}$%
\thanks{$^{1}$Peking University; $^{2}$University of Georgia; $^{3}$ByteDance; $^{4}$Tsinghua University; $^{5}$Chongqing Vocational Institute of Engineering; $^{6}$Infinity Robotics. $^{\dagger}$These authors contributed equally. $^{*}$Corresponding author: Fei Qiao (\texttt{qiaofei@tsinghua.edu.cn}).}}

\begin{document}

\maketitle
\thispagestyle{empty}
\pagestyle{empty}

\begin{abstract}
Per-scene 3D Gaussian Splatting (3DGS) enables high-fidelity rendering, but practical robotic and AR scene capture pipelines often depend on external geometric initialization (e.g., SfM point clouds or depth estimates), which can be slow and brittle in on-site deployment. We present ACEsplat, a fast per-scene optimization framework that reconstructs 3D Gaussian representations from RGB images and camera poses only, without requiring external 3D priors (e.g., precomputed SfM models or supervised depth maps). ACEsplat uses a two-stage pipeline: (1) a self-supervised scene coordinate regression (SCR) module builds an internal geometry prior within 4--5 minutes; (2) SCR features and coordinate priors are fused by a lightweight Gaussian initialization head, followed by per-scene 3DGS optimization. On static-view rendering, ACEsplat achieves 29.11 dB PSNR on Wayspots with real-time SLAM poses and 33.20 dB on Cambridge Landmarks with SfM-refined poses. On RealEstate10K sparse-view novel view synthesis, it achieves competitive image fidelity under a challenging 2-view setting. ACEsplat completes scene-specific SCR mapping and 3DGS reconstruction within 15--25 minutes on a single GPU, making it a practical RGB+pose-only solution for rapid scene setup in robotics and mixed-reality applications.
\end{abstract}

\section{Introduction}
\label{sec:intro}

3D Gaussian Splatting (3DGS)~\cite{kerbl20233d} has become an effective representation for real-time rendering and novel view synthesis, with strong visual fidelity and efficient rendering for VR/AR and robotics-related scene capture applications~\cite{azuma1997survey,jerald2015vr,yuen2011augmented}.
In many practical settings (e.g., on-site mapping, mixed-reality content capture, and robotic field deployment), users only need to rapidly reconstruct a local scene or a small set of viewpoints rather than a complete large-scale environment.
This makes fast static-view rendering and sparse-view novel view synthesis especially important for mixed reality, virtual exhibitions, and remote collaboration~\cite{sakashita2024sharednerf,flavian2019impact,foo2008online}.

Despite its rendering quality, deploying 3DGS to a new scene still requires nontrivial per-scene setup.
High-quality pipelines often rely on external geometric initialization (e.g., SfM point clouds or depth estimates), and the corresponding preprocessing can be time-consuming or brittle under limited views, low texture, or motion blur~\cite{schonberger2016structure,snavely2006photo,wu2011visualsfm}.
In particular, many SfM-initialized per-scene pipelines become unreliable in extreme sparse-view settings (e.g., 2-view novel view synthesis), where stable geometric initialization is difficult.
Recent amortized/feed-forward 3DGS methods~\cite{chen2024mvsplat,charatan2024pixelsplat,du2023learning,suhail2022generalizable} reduce per-scene optimization by learning strong priors from large-scale data, but typically require expensive offline training, often depend on depth or geometric annotations, and are less attractive for rapid on-robot or on-site deployment.

These observations motivate a complementary operating regime: a fast \emph{per-scene} pipeline that avoids external geometric priors and massive pre-training while retaining scene-specific optimization.
To this end, we propose \emph{ACEsplat}, a two-stage per-scene optimization framework that reconstructs scene point clouds and 3D Gaussian representations from \emph{RGB images and camera poses only}.
First, ACEsplat uses a self-supervised Scene Coordinate Regression (SCR) stage to build an internal geometry prior on the fly.
Second, it freezes the SCR network and fuses SCR features and SCR-derived coordinate priors with a lightweight Gaussian initialization head, followed by standard per-scene 3DGS optimization.

We evaluate \emph{ACEsplat} on datasets spanning practical AR/robotics-oriented scene capture and sparse-view synthesis.
On Wayspots~\cite{brachmann2023ace,arnold2022mapfree} with real-time SLAM poses, ACEsplat achieves 29.11 dB PSNR.
On Cambridge Landmarks~\cite{kendall2015posenet} with SfM-refined poses, it reaches 33.20 dB PSNR.
On RealEstate10K~\cite{zhou2018stereo} under a challenging 2-view setting (where SfM-based initialization is often unreliable), ACEsplat achieves competitive sparse-view novel view synthesis performance (26.11 dB PSNR) compared with state-of-the-art amortized models.
Importantly, ACEsplat completes SCR mapping and per-scene 3DGS reconstruction within 15--25 minutes on a single GPU, making it substantially more practical for rapid robotic deployment and on-site adaptation than heavy feed-forward pipelines that rely on large-scale pre-training.

Our contributions are summarized as follows.
\textbf{(1)} We propose \emph{ACEsplat}, a fast two-stage \emph{per-scene} 3DGS reconstruction pipeline using only RGB images and camera poses, without external depth supervision or external geometric priors.
\textbf{(2)} We introduce an SCR-initialized Gaussian construction strategy that uses self-supervised scene-coordinate priors and SCR features to improve 3DGS initialization for subsequent per-scene optimization.
\textbf{(3)} We provide extensive experiments on Wayspots, Cambridge Landmarks, and RealEstate10K, including ablations and runtime--quality analysis, demonstrating competitive rendering quality with strong deployment efficiency in AR/robotics-oriented settings.

\begin{figure*}[tbp]
	\centering
	\mbox{} \hfill
	\includegraphics[width=\linewidth]{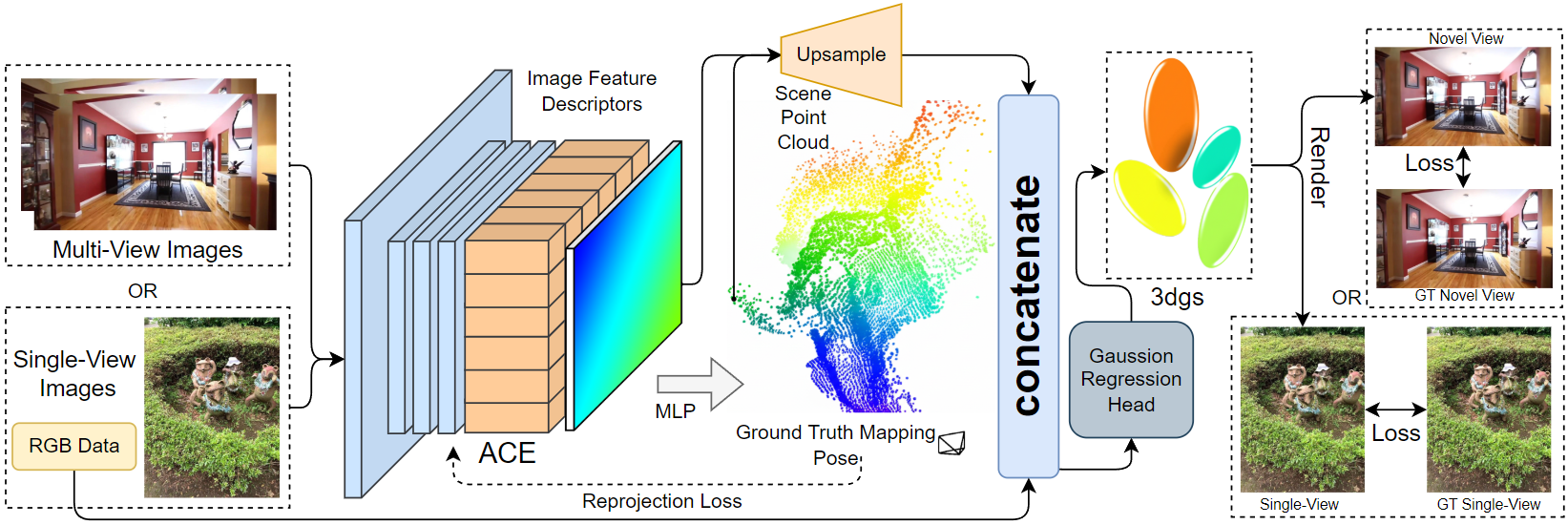}
	\vspace{-10pt}
	\hfill \mbox{}
	\caption{\textbf{Overview of the ACEsplat pipeline.}
		Given RGB images and camera poses, ACEsplat first uses an ACE-based self-supervised scene coordinate regression (SCR) module to produce an \emph{SCR-derived internal geometry prior} (scene coordinates / point cloud) together with image feature descriptors.
		These geometry priors and features are concatenated and fed into a Gaussian attribute initialization head to initialize 3D Gaussian parameters.
		The initialized Gaussians are then refined by per-scene 3DGS optimization under rendering loss for two settings: (i) static-view reconstruction for VR/AR rendering, and (ii) sparse-view novel view synthesis.}
	\label{fig2}
	\vspace{-17pt}
\end{figure*}

\section{Related Work}
\label{sec:related}

\subsection{Visual Relocalization and Scene Coordinate Regression}
Early visual relocalization methods mainly relied on local feature matching and geometric verification.
Recent learning-based approaches introduced point-cloud or point-map prediction paradigms, where an image is mapped to 3D geometry representations using neural networks~\cite{qi2017pointnet}.
More recently, foundation-style geometry models such as Dust3R~\cite{wang2024dust3r} and VGGT~\cite{wang2025vggt} have shown strong structure recovery performance by jointly reasoning about geometry and poses.
While these models are powerful, they are typically designed for large-scale training and high-capacity inference, which may be less suitable for fast per-scene deployment in resource-constrained settings.

A complementary line of work is scene coordinate regression (SCR), which directly predicts 3D scene coordinates from image pixels and avoids explicit feature matching~\cite{shotton2013scene,brachmann2017dsac,brachmann2021visual}.
Recent Accelerated Coordinate Encoding (ACE)~\cite{brachmann2023ace} significantly improves SCR efficiency and enables scene mapping within minutes.
In this work, we build on ACE as a self-supervised SCR module and repurpose its scene-coordinate predictions and feature representations as \emph{internal geometric priors} for 3DGS initialization.
This differs from prior SCR work, which primarily targets camera relocalization rather than scene rendering.

\subsection{3DGS Initialization, Feed-Forward Prediction, and Fast Scene Reconstruction}
3D Gaussian Splatting (3DGS)~\cite{kerbl20233d} enables high-quality real-time rendering, but practical deployment in a new scene still depends on reliable geometric initialization and scene-specific optimization.
Conventional pipelines often rely on SfM point clouds, which can be time-consuming or unstable under sparse views, low-texture regions, or motion blur~\cite{schonberger2016structure,snavely2006photo,wu2011visualsfm}.
Monocular depth estimation~\cite{bhat2023zoedepth,yang2024depth} provides an alternative initialization signal, but depth ambiguity and cross-view inconsistency may limit its effectiveness for stable multi-view Gaussian optimization.

To reduce per-scene optimization, recent feed-forward or amortized 3DGS methods learn geometric and appearance priors from large-scale datasets.
Representative methods include MVSplat~\cite{chen2024mvsplat}, which uses cost volumes for multi-view consistency, and DepthSplat~\cite{geiger2024depthsplat}, which combines multi-view reasoning with monocular depth features.
Other approaches such as LGM~\cite{tang2024lgm}, GRM~\cite{xu2024grm}, and GS-LRM~\cite{zhang2024gs} rely on large-scale training to directly predict Gaussian representations.
A complementary line removes the SfM dependency by jointly optimizing camera poses and Gaussians (e.g., COLMAP-free 3DGS~\cite{fu2024colmapfree}) or by predicting Gaussians from unposed images (e.g., NoPoSplat~\cite{ye2024noposplat}).
These methods are effective, but pose-free joint optimization can be sensitive to initialization under wide baselines or low overlap, and feed-forward variants still rely on substantial offline training.
ACEsplat targets a complementary operating point: a fast \emph{per-scene} pipeline that, given RGB images and camera poses (e.g., from on-device SLAM), replaces external SfM/depth initialization with an internal SCR-based geometry prior fused with SCR features for 3DGS initialization.
This design bridges SCR-based geometry estimation and 3DGS reconstruction for rapid on-site scene setup under sparse-view capture conditions.

\section{Method}
\label{sec:method}
We propose \emph{ACEsplat}, a fast \emph{per-scene} two-stage pipeline for static-view rendering and sparse-view novel view synthesis from RGB images and known camera poses.
Given $N$ scene images $\{I_i\}_{i=1}^{N}$, $I_i \in \mathbb{R}^{H \times W \times 3}$, with camera intrinsics and extrinsics, our goal is to initialize and optimize a scene-level 3D Gaussian representation.
Unlike amortized feed-forward methods that rely on extensive offline pre-training, ACEsplat performs fast scene-specific optimization by replacing \emph{external} geometric initialization (e.g., SfM point clouds or supervised depth maps) with an \emph{internal} geometry prior from self-supervised Scene Coordinate Regression (SCR).
As shown in Fig.~\ref{fig2}, ACEsplat contains two stages: an ACE-based self-supervised SCR module that predicts scene-coordinate priors for the target scene, followed by a lightweight Gaussian initialization head that fuses SCR features and SCR-derived coordinate priors to initialize Gaussian attributes for subsequent per-scene 3DGS optimization.
The SCR backbone and self-supervised objective follow ACE~\cite{brachmann2023ace}; our novelty lies not in the SCR network itself but in repurposing it as a coupled geometry-and-feature source for 3DGS. Unlike generic dense features (e.g., monocular-depth or ImageNet backbones), SCR features are trained by multi-view reprojection and are thus pixel-aligned, scene-specific, and geometrically consistent with the predicted coordinates, which makes them particularly well suited for initializing co-located Gaussian attributes. Our contributions are threefold: (i) reusing SCR-derived scene coordinates as internal geometric priors for 3DGS initialization, (ii) a dense feature upsampling adapter that lifts low-resolution SCR features to pixel-aligned resolution, and (iii) a Gaussian initialization head integrated with per-scene 3DGS optimization.
\subsection{Problem Setup and Representation}
\label{subsec:problem_setup}

Let $\mathcal{S}$ denote the target scene.
ACEsplat constructs and optimizes a scene Gaussian representation
\begin{equation}
	\mathcal{G}=\{(\boldsymbol{\mu}_k,\alpha_k,\boldsymbol{\psi}_k,\mathbf{c}_k)\}_{k=1}^{M},
\end{equation}
where $\boldsymbol{\mu}_k \in \mathbb{R}^{3}$ is the Gaussian center, $\alpha_k$ is opacity, $\boldsymbol{\psi}_k$ denotes the Gaussian shape parameterization (e.g., scale/rotation or covariance-related parameters), and $\mathbf{c}_k$ denotes appearance parameters (e.g., spherical harmonics coefficients).

ACEsplat uses \emph{pixel-aligned Gaussian prediction} for initialization: for each view it predicts a dense coordinate-prior map and feature map at image resolution to instantiate a view-associated Gaussian set, and the merged set is then refined by standard 3DGS optimization (with lifecycle management) rather than used directly as the final representation.

\subsection{ACE-based SCR Geometry Prior}
\label{subsec:scr_prior}

Our first stage builds a scene-specific geometric prior using self-supervised Scene Coordinate Regression (SCR).
We adopt ACE~\cite{brachmann2023ace} as the SCR backbone and training paradigm.
This component is inherited/adapted from ACE and serves as a geometry initializer in ACEsplat.

Given an image location $\mathbf{x}$ (or a local patch centered at $\mathbf{x}$), the SCR network predicts the corresponding 3D scene coordinate:
\begin{equation}
	\mathbf{z} = \mathcal{F}_{\theta}(\mathbf{x}),
	\label{eq:scr_mapping}
\end{equation}
where $\mathcal{F}_{\theta}$ denotes the SCR network parameterized by $\theta$.

Following ACE~\cite{brachmann2023ace}, the SCR model is trained with a reprojection-based robust geometric loss using known camera poses, without requiring ground-truth depth or 3D coordinate annotations.
Compared with explicit feature matching + triangulation pipelines, this implicit coordinate mapping formulation can provide a more stable scene-specific geometric prior in challenging sparse-view conditions.

Let $\mathbf{x}_{j}^{(i)}$ be a pixel (or sampled image location) in image $I_i$, and let $\mathbf{z}_{j}^{(i)}=\mathcal{F}_{\theta}(\mathbf{x}_{j}^{(i)})$ be the predicted scene coordinate.
The SCR optimization objective is:
\begin{equation}
	\min_{\theta}\sum_{i=1}^{N}\sum_{j}\ell_{\pi}\!\left(\mathbf{x}_{j}^{(i)}, \mathbf{z}_{j}^{(i)}, \mathbf{T}_{i}\right),
	\label{eq:scr_obj}
\end{equation}
where $\mathbf{T}_{i}$ is the camera pose of $I_i$, and $\ell_{\pi}$ is the DSAC$^*$-style robust reprojection loss~\cite{brachmann2021visual}.

Here $\ell_{\pi}$ is the DSAC$^*$ robust reprojection loss: valid predictions (those satisfying depth and reprojection constraints) are penalized by a $\tanh$-clamped reprojection error under a circular curriculum on the inlier threshold $\tau(t)$, while invalid predictions fall back to a fixed scene-coordinate target (we refer to~\cite{brachmann2021visual,brachmann2023ace} for the exact formulation).

After SCR training, we freeze the SCR network and use its outputs to construct an internal geometry prior for 3DGS initialization.
For each input view $I_i$, we obtain a scene-coordinate prior map
\begin{equation}
	\mathbf{P}_i \in \mathbb{R}^{H \times W \times 3},
\end{equation}
and a validity mask
\begin{equation}
	\mathbf{M}_i \in \{0,1\}^{H \times W},
\end{equation}
where $\mathbf{M}_i(u,v)=1$ indicates a valid coordinate prior after geometric validity checks (and optional interpolation-validity checks).
Only valid pixels are used to instantiate Gaussians in Stage 2.

\subsection{Dense Feature Upsampling Adapter}
\label{subsec:feature_adapter}

In addition to scene coordinates, ACEsplat reuses intermediate feature maps from the ACE SCR backbone as geometry-aware image features.
However, the native SCR feature resolution is lower than the input image resolution and is not directly suitable for pixel-aligned Gaussian attribute initialization.

To address this, we introduce a lightweight dense feature upsampling adapter (ours), which progressively upsamples SCR feature maps to image resolution.
Inspired by U-Net/FPN-style dense prediction designs~\cite{ronneberger2015u,lin2017feature}, the adapter combines bilinear interpolation and convolutional layers to produce a dense feature map aligned with the input image:
\begin{equation}
	\mathbf{F}_i \in \mathbb{R}^{H \times W \times C}.
\end{equation}
Compared with transposed-convolution alternatives~\cite{gao2019pixel,sugawara2019checkerboard}, this design is intended to avoid checkerboard artifacts and keep training stable.
Concretely, the adapter is a shallow module (a few bilinear-upsampling and $3\times3$ convolution blocks) applied once per view prior to optimization, and is small relative to the main per-scene 3DGS optimization.

\subsection{Gaussian Attribute Initialization Head}
\label{subsec:gaussian_init}

Given the pixel-aligned SCR feature map $\mathbf{F}_i$ and SCR-derived coordinate prior map $\mathbf{P}_i$ for view $I_i$, we concatenate them along the channel dimension:
\begin{equation}
	\mathbf{X}_i = [\mathbf{F}_i \,\|\, \mathbf{P}_i] \in \mathbb{R}^{H \times W \times (C+3)}.
	\label{eq:concat_fp}
\end{equation}

We then use a lightweight fully convolutional Gaussian initialization head $\mathcal{H}_{\phi}$ to predict Gaussian attributes other than the center:
\begin{equation}
	[\boldsymbol{\alpha}_i,\mathbf{A}_i,\boldsymbol{\Psi}_i] = \mathcal{H}_{\phi}(\mathbf{X}_i),
	\label{eq:gaussian_head}
\end{equation}
where $\boldsymbol{\alpha}_i$ denotes opacity predictions, $\mathbf{A}_i$ denotes appearance parameter predictions (e.g., spherical harmonics coefficients), and $\boldsymbol{\Psi}_i$ denotes Gaussian shape parameter predictions (e.g., scale/rotation or covariance-related parameters).

The Gaussian centers are initialized directly from SCR-derived coordinates:
\begin{equation}
	\boldsymbol{\mu}_{i}(u,v) = \mathbf{P}_{i}(u,v), \quad \text{for } \mathbf{M}_i(u,v)=1.
	\label{eq:center_from_scr}
\end{equation}

In practice, SCR coordinate predictions may be sparse or lower-resolution depending on the implementation and sampling strategy.
We therefore align coordinate priors to image resolution (via interpolation if needed) and retain only valid pixels using $\mathbf{M}_i$, ensuring dimensional consistency among the input image, coordinate prior map, and dense feature map while reducing noisy initialization.

Unlike SCR, where predictions can be made independently for sampled locations, rendering quality depends on local spatial structure.
We therefore use a fully convolutional design to explicitly model neighborhood dependencies when predicting Gaussian attributes.

\paragraph{View-wise Gaussian instantiation and scene-level merging.}
For each input view $I_i$, we instantiate a view-associated Gaussian set from valid pixels:

\begin{equation}
	\small
	\mathcal{G}_i^{(0)}\!=\!
	\left\{
	\!(\boldsymbol{\mu}_{i}(u,v),\alpha_{i}(u,v),\boldsymbol{\psi}_{i}(u,v),\mathbf{a}_{i}(u,v))
	\;\middle|\;
	\mathbf{M}_i(u,v)\!=\!1
	\right\}\!,
	\label{eq:viewwise_init}
\end{equation}

where $\alpha_i(u,v)$, $\boldsymbol{\psi}_i(u,v)$, and $\mathbf{a}_i(u,v)$ are sampled from the predicted maps $\boldsymbol{\alpha}_i$, $\boldsymbol{\Psi}_i$, and $\mathbf{A}_i$ at pixel $(u,v)$.
The initial scene Gaussian set is formed by merging all view-associated sets:
\begin{equation}
	\mathcal{G}^{(0)} = \bigcup_{i=1}^{N} \mathcal{G}_i^{(0)}.
	\label{eq:scene_init_merge}
\end{equation}
This initialization is intentionally over-complete and is later compacted by Gaussian lifecycle operations during optimization.

\subsection{Per-scene 3DGS Optimization and Gaussian Lifecycle}
\label{subsec:gs_optimization}

The pixel-aligned Gaussian prediction described above is used for initialization, while the final representation is obtained through rendering-based per-scene optimization.
Let $\mathbf{I}_{\text{render}}$ be the rendered image from the current Gaussian set and $\mathbf{I}_{\text{gt}}$ be the target image.
We optimize a photometric reconstruction loss:
\begin{equation}
	\mathcal{L}_{\text{gs}}
	=
	(1-\lambda)\,\mathcal{L}_{\text{MSE}}(\mathbf{I}_{\text{render}}, \mathbf{I}_{\text{gt}})
	+
	\lambda\,\mathcal{L}_{\text{D-SSIM}}(\mathbf{I}_{\text{render}}, \mathbf{I}_{\text{gt}}),
	\label{eq:photometric_loss}
\end{equation}
where $\mathcal{L}_{\text{MSE}}$ and $\mathcal{L}_{\text{D-SSIM}}$ are the pixel-wise MSE and differentiable SSIM losses, respectively.

In Stage 2, the SCR network is frozen and the Gaussian initialization head is optimized \emph{per scene} (not as a cross-scene amortized predictor); we then jointly optimize the head parameters and the scene Gaussian representation under rendering supervision within the same per-scene budget.
Because the pixel-aligned initialization is intentionally over-complete, we apply Gaussian lifecycle management (pruning low-contribution Gaussians and optional densification/splitting) during optimization to converge to a compact representation; exact settings are given in Sec.~\ref{sec:exp}.

\subsection{Task-Specific Training Protocols}
\label{subsec:task_protocols}

We use the same SCR-initialized Gaussian construction in both settings, differing only in input views and supervision.
For \emph{static-view VR/AR rendering}, a single input view initializes and optimizes a scene representation anchored by SCR-derived geometry for high-fidelity reconstruction of the captured view.
For \emph{sparse-view novel view synthesis}, two input views initialize the representation and held-out target views supervise synthesis; relative to optimization from weak initialization, SCR-based anchoring improves cross-view consistency in the 2-view regime.
In both cases we first optimize SCR, then freeze it and jointly optimize the initialization head and per-scene 3DGS, with schedules reported in Sec.~\ref{sec:exp}.

\section{Experiments}
\label{sec:exp}

\subsection{Experimental Setup}
\label{subsec:exp_setup}

\subsubsection{Implementation details and runtime accounting}
We implement ACEsplat in PyTorch~\cite{paszke2017automatic} based on the public ACE codebase~\cite{brachmann2023ace}.
All experiments are conducted on a single NVIDIA A40 GPU.
Stage 1 trains the ACE-based SCR module using ACE default settings and takes approximately 4--5 minutes per scene.
Stage 2 optimizes the Gaussian initialization head together with per-scene 3DGS optimization using AdamW~\cite{loshchilov2017decoupled} (learning rate $2\times10^{-4}$ to $2\times10^{-3}$, one-cycle schedule, mixed precision).
For static-view rendering, Stage 2 runs for 8 epochs (batch size 16, $\sim$10 minutes).
For sparse-view novel view synthesis, Stage 2 runs for 16 epochs (batch size 10, $\sim$20 minutes).
Unless otherwise noted, we report runtime as Stage 1 (SCR mapping) + Stage 2 (Gaussian initialization + per-scene 3DGS optimization).

\subsubsection{3DGS optimization details}
Unless otherwise stated, we adopt the default 3DGS lifecycle settings~\cite{kerbl20233d}: adaptive densification runs from iteration 500 to 15{,}000 with a densification interval of 100 and a positional-gradient threshold of $2\times10^{-4}$, opacity is reset every 3{,}000 iterations, and low-contribution Gaussians (opacity below 0.005) are pruned.
Because the pixel-aligned initialization is intentionally over-complete, this pruning is essential for converging to a compact model; final per-scene Gaussian counts and image resolutions are reported in the supplementary material.

\subsubsection{Datasets and protocols}
We evaluate ACEsplat on Wayspots~\cite{brachmann2023ace,arnold2022mapfree} (static-view rendering with real-time SLAM poses), Cambridge Landmarks~\cite{kendall2015posenet} (static-view rendering with SfM-refined poses), and RealEstate10K~\cite{zhou2018stereo} (sparse-view novel view synthesis at $256\times256$).
For Wayspots, we use the dataset's device-recorded real-time visual--inertial poses (on-device ARKit-style tracking~\cite{oufqir2020arkit}) rather than offline SfM~\cite{arnold2022mapfree}, which is precisely the noisy-pose regime our SLAM-pose setting targets; Cambridge Landmarks instead provides SfM-refined poses.
For RealEstate10K, we evaluate on 500 scenes randomly sampled from the official test split with a fixed seed (42).
We filter out sequences with insufficient camera translation or fewer than 30 valid frames to ensure meaningful wide-baseline evaluation.
For the 2-view protocol, we extract a local segment from each filtered scene, use the first and last frames as sparse input views, and uniformly sample intermediate frames as target views for metric computation.

\subsubsection{Evaluation metrics}
We report PSNR, SSIM~\cite{1284395}, and LPIPS~\cite{zhang2018unreasonable} by comparing rendered outputs with reference images.
For Wayspots, we additionally report ACE localization accuracy (percentage of frames with pose error below \SI{10}{\centi\metre} translation and \SI{5}{\degree} rotation), since SCR quality directly affects the geometric consistency of the downstream Gaussian initialization.

\subsubsection{Baselines}
We compare ACEsplat with a conventional SfM+3DGS per-scene pipeline for static-view rendering, and with representative sparse-view novel view synthesis methods for the 2-view RealEstate10K setting.
For RealEstate10K (2-view), we do not include a vanilla SfM+3DGS baseline as a primary comparison because SfM-based initialization is often unreliable under extreme two-view settings (limited overlap / challenging baselines), which frequently leads to unstable or degenerate Gaussian optimization.
Consequently, amortized/feed-forward sparse-view methods constitute the most meaningful baselines for this setting.
We further emphasize that ACEsplat and these baselines operate under different compute regimes: ACEsplat is a fast \emph{per-scene} optimization pipeline, whereas feed-forward baselines rely on substantial offline training.
Therefore, the training-time and hardware entries in Tab.~\ref{tab3} are reported as contextual compute references that characterize the deployment regime, and should not be read as a controlled per-scene speedup comparison.

\begin{figure}[htbp]
	\centering
	\includegraphics[width=\linewidth]{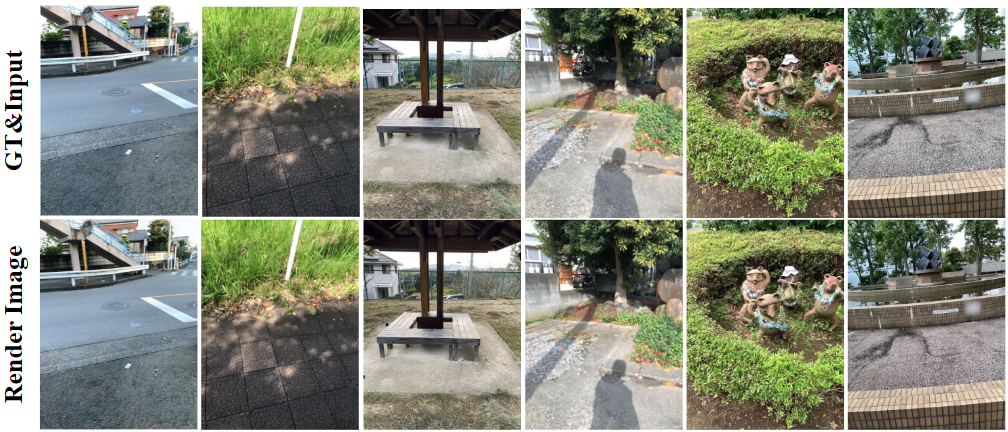}
	\vspace{-10pt}
	\caption{\label{fig_wayspots}
		\textbf{Qualitative results on the Wayspots dataset with real-time SLAM poses.}
		Given single RGB inputs and SLAM poses, ACEsplat reconstructs high-fidelity static views across diverse outdoor scenes.}
	\vspace{-8pt}
\end{figure}

\begin{figure}[htbp]
	\centering
	\includegraphics[width=\linewidth]{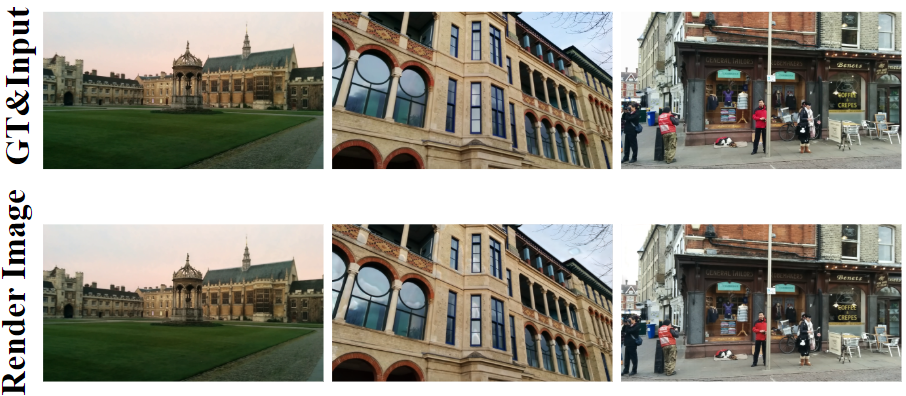}
	\vspace{-15pt}
	\caption{\label{fig3}
		\textbf{Qualitative results on the Cambridge Landmarks dataset.}
		ACEsplat achieves strong static-view rendering quality on large-scale scenes using SCR-derived geometric priors and SfM-refined poses.}
	\vspace{-15pt}
\end{figure}

\begin{figure}[t]
	\centering
	\scalebox{1}{\includegraphics[width=0.7\linewidth]{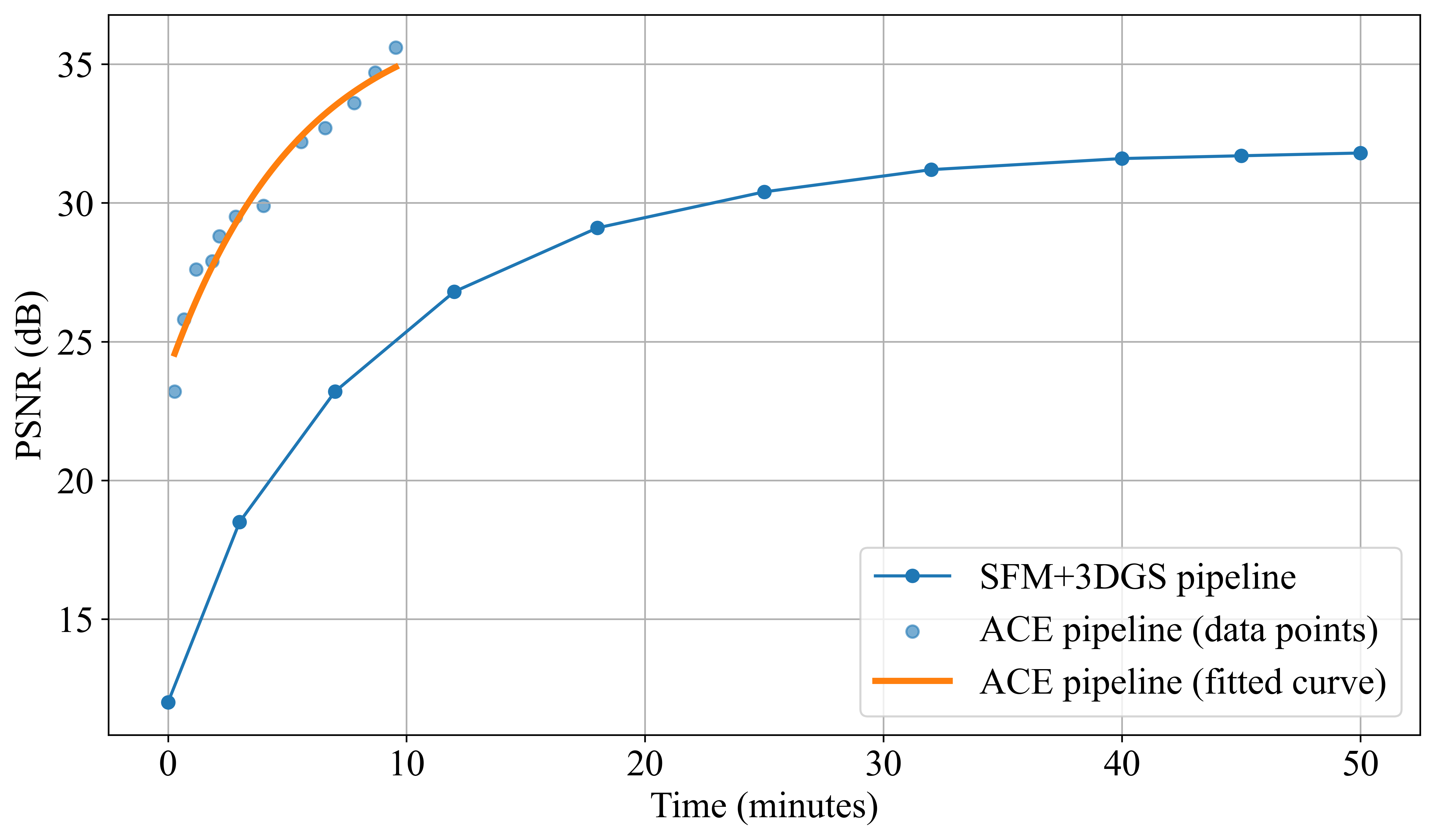}}
	\vspace{-15pt}
	\caption{
		\textbf{Runtime vs.\ rendering quality on Cambridge Landmarks (ACEsplat vs.\ SfM+3DGS).}
		At every runtime budget ACEsplat attains higher PSNR than the SfM+3DGS pipeline and reaches strong quality within a few minutes of per-scene optimization, avoiding the lengthy SfM feature-extraction and matching stages.
		The plotted \emph{SfM+3DGS} configuration and runtime accounting are detailed in Sec.~\ref{subsec:exp_setup}.}
	\label{fig4}
	\vspace{-15pt}
\end{figure}

\subsection{Static-View Rendering}
\label{subsec:static_view_rendering}

\subsubsection{Wayspots: Rendering with Real-time SLAM Poses}
Real-time SLAM poses are often noisy in challenging outdoor environments, which directly affects reprojection-based SCR training and the quality of downstream 3DGS initialization.
To quantify this effect, we report ACE localization accuracy together with rendering metrics in Tab.~\ref{tab1}.
We observe that higher localization accuracy generally correlates with better rendering quality on Wayspots: more accurate SCR geometry priors lead to more spatially consistent Gaussian initialization and improved rendering fidelity.
Conversely, accumulated SLAM pose errors can introduce geometric inconsistencies in the SCR stage and degrade the final rendering quality.
We emphasize that the reported accuracy is the fraction of frames within a strict \SI{10}{\centi\metre}/\SI{5}{\degree} threshold, so a low or zero value (e.g., \emph{wayspots\_statue}) does not imply that SCR has failed: the regressed coordinates can remain globally consistent enough to anchor a usable Gaussian initialization, with per-frame pose error merely exceeding the strict threshold, which is why rendering still reaches 23.30 dB.
Noisy poses do affect the reprojection-based SCR stage, but its robust loss and curriculum tolerate moderate SLAM noise rather than requiring exact poses.
Despite these challenges, ACEsplat achieves an average PSNR of 29.11 dB across ten outdoor Wayspots scenes (Tab.~\ref{tab1}), with qualitative examples shown in Fig.~\ref{fig_wayspots}.
From a runtime perspective, ACEsplat requires approximately 5 minutes for SCR mapping (Stage 1) and 10 minutes for Stage-2 optimization on this task, yielding a total per-scene runtime of about 15 minutes on a single GPU.
This runtime profile is attractive for on-site AR-style deployment where rapid adaptation to a newly captured scene is important.

\subsubsection{Cambridge Landmarks: Large-scale Static-View Rendering}
We further evaluate ACEsplat on Cambridge Landmarks with SfM-refined poses to assess scalability in larger outdoor scenes.
As shown in Tab.~\ref{tab2}, ACEsplat achieves an average PSNR of 33.20 dB.
Although ACE may exhibit local ambiguities in large scenes with repetitive structures (e.g., repeated windows or facade patterns), the resulting SCR-derived geometry priors remain sufficiently consistent to support high-quality downstream 3DGS optimization.
Qualitative examples are shown in Fig.~\ref{fig3}.
This behavior highlights an important property of ACEsplat: the SCR stage does not need to recover perfect geometry to be useful.
Instead, it provides a fast and reasonably consistent internal geometry prior that substantially improves per-scene 3DGS initialization.
Rather than adding a separate SfM+3DGS column to Tab.~\ref{tab1} and Tab.~\ref{tab2}, we report the head-to-head comparison as a runtime-versus-PSNR curve in Fig.~\ref{fig4}, since the two pipelines reach comparable final quality and differ mainly in how quickly they get there.
By bypassing the lengthy feature-extraction and multi-view matching stages of SfM, ACEsplat attains higher PSNR than SfM+3DGS at every runtime budget and converges within a few minutes.

\begin{table}[htbp]
	\centering
	\setlength{\tabcolsep}{3pt} 
	\renewcommand{\arraystretch}{0.9} 
	\setlength{\aboverulesep}{0pt}
	\setlength{\belowrulesep}{0pt}
	\setlength{\extrarowheight}{0pt}
	\small
	\caption{\textbf{Static-view rendering results on the Wayspots dataset~\cite{brachmann2023ace,arnold2022mapfree}.}}
	\vspace{-10pt}
	\resizebox{\columnwidth}{!}{
		\begin{tabular}{lcccc}
			\toprule
			Scene                        & \makecell{Accuracy (\%)} & \makecell{PSNR\textuparrow} & \makecell{SSIM\textuparrow} & \makecell{LPIPS\textdownarrow} \\
			\midrule
			\emph{wayspots\_bears}       & 80.7                     & 34.17                       & 0.9844                      & 0.0339                         \\
			\emph{wayspots\_cubes}       & 97.0                     & 30.99                       & 0.9789                      & 0.0539                         \\
			\emph{wayspots\_inscription} & 49.0                     & 29.39                       & 0.9669                      & 0.0593                         \\
			\emph{wayspots\_map}         & 56.5                     & 34.89                       & 0.9858                      & 0.0319                         \\
			\emph{wayspots\_squarebench} & 66.7                     & 30.92                       & 0.9622                      & 0.0930                         \\
			\emph{wayspots\_therock}     & 100.0                    & 28.17                       & 0.8706                      & 0.2059                         \\
			\emph{wayspots\_lawn}        & 35.8                     & 22.68                       & 0.9154                      & 0.1242                         \\
			\emph{wayspots\_statue}      & 0.0                      & 23.30                       & 0.8797                      & 0.1909                         \\
			\emph{wayspots\_tendrils}    & 34.9                     & 27.39                       & 0.9520                      & 0.1035                         \\
			\emph{wayspots\_wintersign}  & 1.0                      & 29.24                       & 0.8998                      & 0.1822                         \\
			\midrule
			\textbf{Average}             & \textbf{52.16}           & \textbf{29.11}              & \textbf{0.9396}             & \textbf{0.1079}                \\
			\bottomrule
		\end{tabular}
	}
	\label{tab1}
	\vspace{-10pt}
\end{table}

\begin{table}[htbp]
	\centering
	\setlength{\tabcolsep}{3pt}
	\renewcommand{\arraystretch}{0.9}
	\setlength{\aboverulesep}{0pt}
	\setlength{\belowrulesep}{0pt}
	\setlength{\extrarowheight}{0pt}
	\small
	\caption{\textbf{Static-view rendering results on Cambridge Landmarks.}
		Pose statistics report the average supervision pose error (translation / rotation).}
	\vspace{-10pt}
	\resizebox{\columnwidth}{!}{
		\begin{tabular}{lccccc}
			\toprule
			Scene               & \makecell{Pose Error                                  \\ (\si{\centi\metre} / $^\circ$)} & PSNR (dB)\textuparrow & SSIM\textuparrow & LPIPS\textdownarrow & Time\textdownarrow \\
			\midrule
			{\emph{Hospital}}   & 31/0.6               & 31.82 & 0.9679 & 0.0769 & 599s \\
			{\emph{King}}       & 28/0.4               & 31.88 & 0.9598 & 0.1153 & 607s \\
			{\emph{GreatCourt}} & 43/0.2               & 36.57 & 0.9754 & 0.0933 & 603s \\
			{\emph{Shop}}       & 5/0.3                & 31.80 & 0.9609 & 0.0757 & 597s \\
			{\emph{StMary}}     & 18/0.6               & 33.92 & 0.9744 & 0.0749 & 609s \\
			\midrule
			Average             & 25/0.4               & 33.20 & 0.9677 & 0.0872 & 603s \\
			\bottomrule
		\end{tabular}
	}
	\label{tab2}
	\vspace{-15pt}
\end{table}

\subsection{Sparse-View Novel View Synthesis}
\label{subsec:nvs}
\begin{figure}[tbp]
	\centering
	\mbox{} \hfill
	\includegraphics[width=\linewidth]{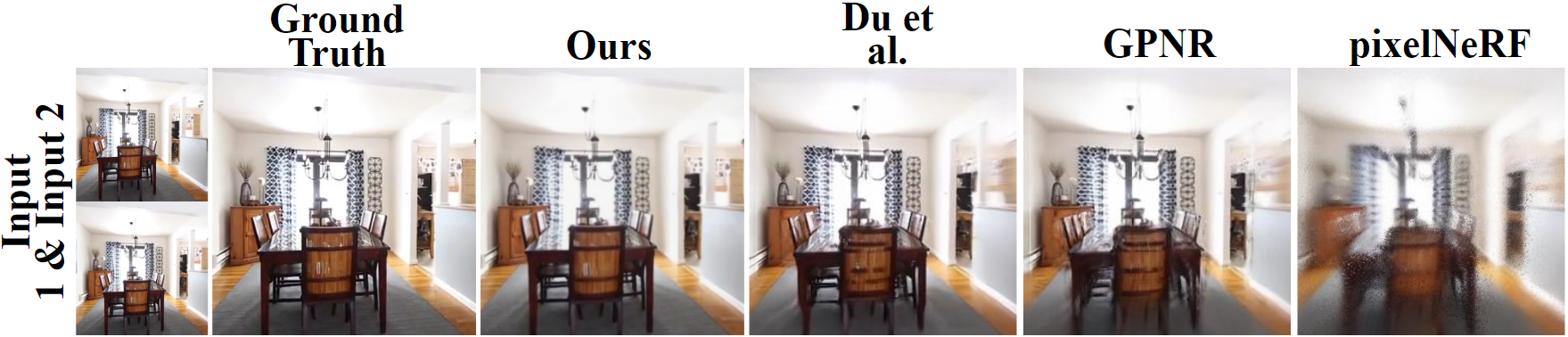}
	\hfill \mbox{}
	\vspace{-10pt}
	\caption{\textbf{Qualitative comparison on RealEstate10K (sparse-view novel view synthesis).}
		Given two input views (left), ACEsplat produces competitive novel-view renderings and preserves scene structure more faithfully than several prior sparse-view methods~\cite{du2023learning,suhail2022generalizable,yu2021pixelnerf}.}
	\label{fig5}
	\vspace{-30pt}
\end{figure}

We evaluate ACEsplat on 500 RealEstate10K scenes under a fixed 2-view sparse-view protocol.
Compared with static-view rendering datasets, RealEstate10K has sparser and more diverse scene coverage, making SCR training and geometry initialization more challenging.
In this extreme 2-view regime, where SfM-based initialization is often unreliable for per-scene Gaussian optimization, ACEsplat's internal geometric prior provides effective structural anchoring.
As a result, ACEsplat achieves competitive performance with an average PSNR of 26.11 dB (Tab.~\ref{tab3}) with short per-scene runtime on a single GPU.
Table~\ref{tab3} compares ACEsplat with representative sparse-view methods under a different operating regime: ACEsplat performs fast on-the-fly per-scene optimization, while the baselines are feed-forward models supported by substantial offline training.
The comparison with DepthSplat~\cite{geiger2024depthsplat} highlights this deployment trade-off: DepthSplat achieves higher metrics with explicit depth supervision and large-scale pre-training, whereas ACEsplat provides a practical RGB+pose-only per-scene alternative with competitive fidelity and short single-GPU adaptation time.
Qualitative results in Fig.~\ref{fig5} further show that ACEsplat produces visually competitive novel-view renderings and preserves scene structure more consistently than earlier baselines, including pixelNeRF~\cite{yu2021pixelnerf}, GPNR~\cite{suhail2022generalizable}, and Du et al.~\cite{du2023learning}.

\begin{table}[htbp]
	\centering
	\setlength{\tabcolsep}{4pt}
	\renewcommand{\arraystretch}{0.9}
	\setlength{\aboverulesep}{0pt}
	\setlength{\belowrulesep}{0pt}
	\setlength{\extrarowheight}{0pt}
	\small
	\vspace{-15pt}
	\caption{\textbf{Novel view synthesis results on RealEstate10K.}}
	\vspace{-10pt}
	\resizebox{\columnwidth}{!}{
		\begin{tabular}{lccccc}
			\toprule
			Method                                 & PSNR (dB)\textuparrow & SSIM\textuparrow & LPIPS\textdownarrow & Time\textdownarrow            & GPU Configuration   \\
			\midrule
			DepthSplat~\cite{geiger2024depthsplat} & 27.47                 & 0.889            & 0.114               & \makecell{2 days}             & 4$\times$ GH200     \\
			Ours (ACEsplat)                        & 26.11                 & 0.876            & 0.202               & \makecell[l]{20 min (Stage 2)                       \\25 min total} & 1$\times$ A40 \\
			Du et al.~\cite{du2023learning}        & 24.78                 & 0.820            & 0.213               & \makecell{3 days}             & 4$\times$ V100      \\
			GPNR~\cite{suhail2022generalizable}    & 24.11                 & 0.793            & 0.255               & \makecell{1 day}              & 32$\times$ TPU      \\
			pixelNeRF~\cite{yu2021pixelnerf}       & 20.43                 & 0.589            & 0.550               & \makecell{6 days}             & 1$\times$ Titan RTX \\
			\bottomrule
		\end{tabular}
	}
	\label{tab3}
	\vspace{1pt}
	\begin{minipage}{\columnwidth}
		\footnotesize \emph{Note:} Prior-method time/hardware entries are reported from their papers and are not directly comparable to ACEsplat's per-scene optimization time.
	\end{minipage}
	\vspace{-16pt}
\end{table}

\subsection{Ablation Studies}
\label{subsec:ablation}

\begin{figure}[tbp]
	\centering
	\includegraphics[width=\linewidth]{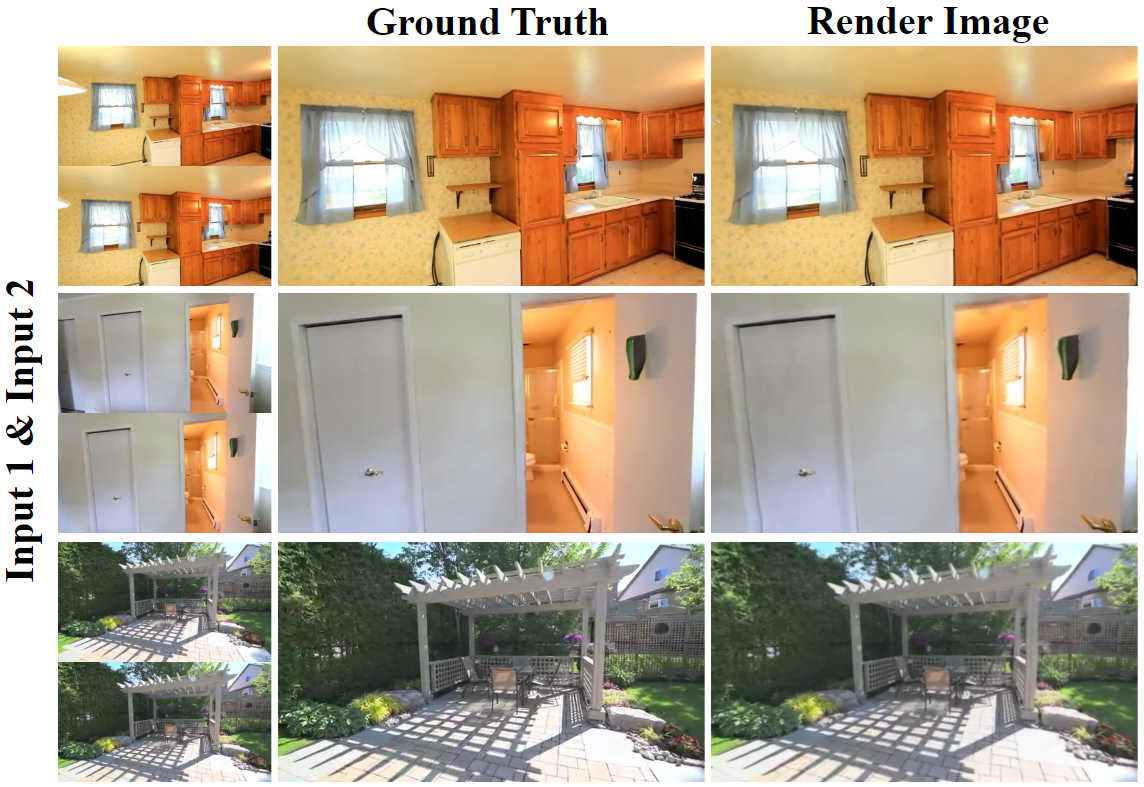}
	\caption{\label{fig6}
		\textbf{High-resolution novel view synthesis on RealEstate10K.}
		ACEsplat maintains fine details and color fidelity on upscaled scenes (input / ground truth / rendered view).}
	\vspace{-20pt}
\end{figure}

\begin{figure}[htbp]
	\centering
	\vspace{-10pt}
	\includegraphics[width=\linewidth]{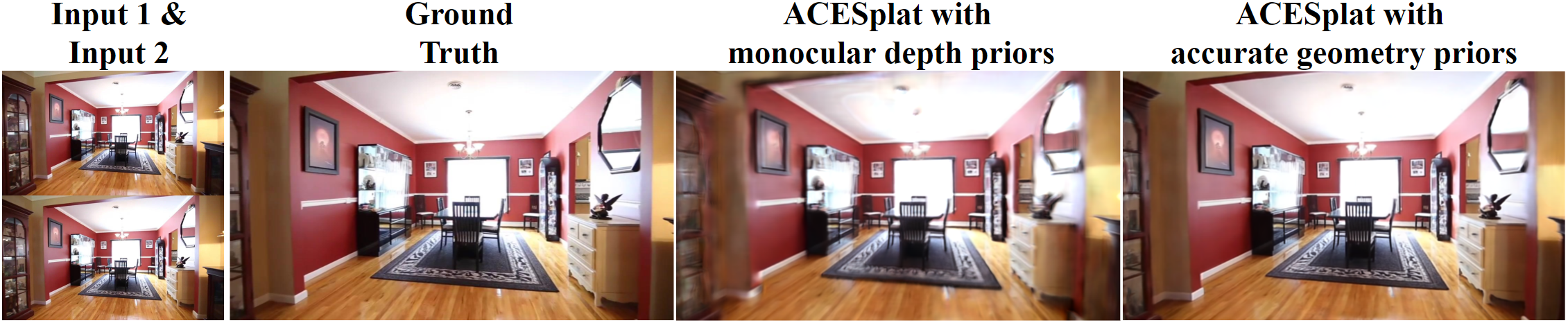}
	\caption{\label{fig7}
		\textbf{Ablation on geometric priors for sparse-view novel view synthesis.}
		ACE-based SCR priors produce more consistent geometry and higher-fidelity renderings than monocular-depth-based priors from ZoeDepth~\cite{bhat2023zoedepth}.}
	\vspace{-0pt}
\end{figure}

\begin{figure}[htbp]
	\centering
	\vspace{-0pt}
	\includegraphics[width=\linewidth]{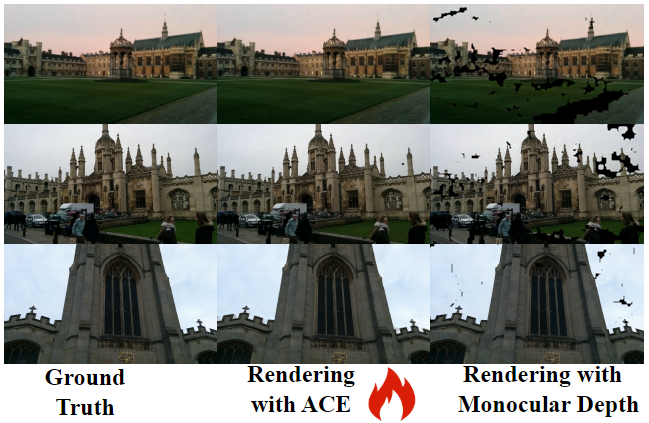}
	\caption{\label{fig8}
		\textbf{Ablation on geometric priors for static-view rendering.}
		Monocular-depth-based priors lead to artifacts (fragmentation and darkening), while SCR-derived priors from ACE produce more stable and faithful renderings.}
	\vspace{-15pt}
\end{figure}

\subsubsection{High-resolution scenarios}
On upscaled RealEstate10K images at $360\times640$, ACEsplat preserves fine details and color consistency (Fig.~\ref{fig6}), indicating that the SCR-initialized Gaussian construction remains effective beyond the default $256\times256$ setting.

\subsubsection{Importance of geometry priors}
We compare ACE-based SCR priors with monocular-depth-based priors from ZoeDepth~\cite{bhat2023zoedepth}.
For the depth baseline, predicted depth maps are unprojected into pseudo scene coordinates and the scene regression branch is trained with a coordinate loss:
\begin{equation}
	\arg\min_{\mathbf{w}} \sum_{i=1}^{N}\sum_{j \in I_i} \left\|\hat{\mathbf{y}}_{ij} - \mathbf{y}_{ij}\right\|^2,
	\label{eq:depth_ablation}
\end{equation}
where $\mathbf{y}_{ij}$ is obtained from ZoeDepth-based monocular depth unprojection.
Due to scale ambiguity and weak cross-view consistency, this baseline yields much lower quality (20.23 dB on RealEstate10K and 27.68 dB on Cambridge Landmarks).
In contrast, ACE-based SCR enforces multi-view geometric consistency through reprojection, producing more coherent coordinate priors for Gaussian initialization.
Qualitative results in Fig.~\ref{fig7} and Fig.~\ref{fig8} further confirm the robustness of SCR-derived priors.
\subsubsection{Effectiveness of the feature description network for Gaussian initialization}
\begin{figure}[htbp]
	\centering
	\includegraphics[width=0.8\linewidth]{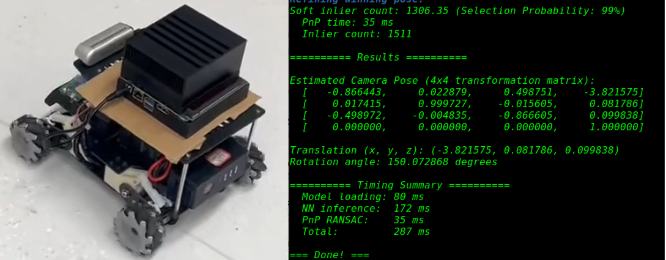}
	\caption{\label{fig_real_robot_demo}
		\textbf{On-robot localization demo.}
		ACE/SCR-based localization component successfully estimates camera pose on a wheeled mobile robot with monocular camera.}
	\vspace{-15pt}
\end{figure}

Replacing the SCR feature pathway with a PointNet~\cite{qi2017pointnet} extractor on ACE-generated point clouds causes unstable optimization, with training quality saturating at a low level (approximately 5.3 dB); this indicates PointNet fails to provide effective dense image-aligned features, supporting the use of spatially coherent SCR backbone features with our dense upsampling adapter.
\subsubsection{On-Robot Localization Demo (Fig.~\ref{fig_real_robot_demo})}
\label{subsec:robot_demo}
As a deployment feasibility check (not a full-pipeline benchmark), we run the ACE/SCR localization component on a wheeled mobile robot with an onboard computer and monocular camera.
In a representative run it estimates the camera pose from 1511 inliers (99\% selection probability) with 287 ms total latency (207 ms excluding one-time model loading: 172 ms inference + 35 ms PnP-RANSAC), supporting the practicality of RGB-based SCR localization for robotics.

\subsection{Discussion and Limitations}
\label{subsec:limitations}
\emph{(i) Spherical harmonics in the 2-view regime.} To limit the SH over-fitting and color shifts that arise in extreme sparse views, ACEsplat relies primarily on low-order SH initialized from RGB and predicted per view from SCR features, and refines them jointly under photometric supervision over the input views (held-out target views are used only for evaluation), which discourages view-inconsistent appearance; a systematic study of SH-degree scheduling is left to future work.
\emph{(ii) SCR failure modes.} Because the SCR prior is learned from reprojection, it can be unreliable in textureless or highly reflective regions where photometric and geometric cues are weak, lowering localization accuracy and rendering quality (e.g., \emph{wayspots\_lawn}/\emph{statue}); per-scene 3DGS optimization partially compensates but does not fully resolve these cases.
\emph{(iii) Fairness of comparisons.} A denser initialization generally speeds up 3DGS and can improve novel-view metrics, so our ablations isolate the prior under a fixed budget (SCR vs.\ monocular-depth priors, and SCR vs.\ PointNet features) rather than only reporting end-to-end timing, and the feed-forward entries in Tab.~\ref{tab3} are contextual rather than controlled speedups.
\emph{(iv) Recent pose-free 3DGS.} We position ACEsplat relative to joint pose--Gaussian and pose-free methods~\cite{fu2024colmapfree,ye2024noposplat} in discussion; a controlled head-to-head benchmark and a standalone latency/memory ablation of the upsampling adapter are promising but beyond the scope of this paper.

\section{Conclusions}
\label{sec:conclusions}
In this paper, we present ACEsplat, a fast per-scene optimization framework for efficient, high-fidelity 3D reconstruction from RGB images and camera poses. By combining a self-supervised Scene Coordinate Regression (SCR) module with a 3D Gaussian Splatting (3DGS) initialization head, ACEsplat achieves robust geometry initialization and high-quality scene reconstruction without supervised depth data or external geometric priors. It performs strongly in AR static-view rendering on the Wayspots and Cambridge Landmarks datasets and achieves competitive image fidelity for sparse-view novel view synthesis on RealEstate10K. ACEsplat completes scene-specific reconstruction in 15--25 minutes on a single GPU, offering a practical RGB+pose-only solution for rapid deployment. These results position ACEsplat as a compelling approach for fast scene reconstruction in AR/VR and robotics-oriented applications.

\section*{Acknowledgment}
The authors acknowledge support from Beijing Natural Science Foundation (L253009), National Natural Science Foundation of China (U25A20489, 62334006), National Science and Technology Major Project Fund of China (2025ZD0215600), National Key Technologies R\&D Program of China (2025YFF1500600), and projects KJQN202503423 and CSTB2025NSCQ-GPX0799; the work of Haohua Que, Tianle Zhu, and Handong Yao was carried out entirely in the United States and did not receive any of this support.


	{\footnotesize
		\bibliographystyle{IEEEtran}
		\bibliography{ref}

@String(CVPR= {IEEE Conf. Comput. Vis. Pattern Recog.})

@String(ICCV= {Int. Conf. Comput. Vis.})

@String(ECCV= {Eur. Conf. Comput. Vis.})

@String(ICLR = {Int. Conf. Learn. Represent.})

@String(VR   = {Vis. Res.})

@String(CVPR  = {CVPR})

@String(ICCV  = {ICCV})

@String(ECCV  = {ECCV})

@String(ICLR  = {ICLR})

@article{yuen2011augmented,
  title={Augmented reality: An overview and five directions for AR in education},
  author={Yuen, Steve Chi-Yin and Yaoyuneyong, Gallayanee and Johnson, Erik},
  journal={Journal of Educational Technology Development and Exchange (JETDE)},
  volume={4},
  number={1},
  pages={11},
  year={2011}
}

@article{azuma1997survey,
  title={A survey of augmented reality},
  author={Azuma, Ronald T},
  journal={Presence: teleoperators \& virtual environments},
  volume={6},
  number={4},
  pages={355--385},
  year={1997},
  publisher={MIT Press One Rogers Street, Cambridge, MA 02142-1209, USA journals-info~…}
}

@book{jerald2015vr,
  title={The VR book: Human-centered design for virtual reality},
  author={Jerald, Jason},
  year={2015},
  publisher={Morgan \& Claypool}
}

@article{flavian2019impact,
  title={The impact of virtual, augmented and mixed reality technologies on the customer experience},
  author={Flavi{\'a}n, Carlos and Ib{\'a}{\~n}ez-S{\'a}nchez, Sergio and Or{\'u}s, Carlos},
  journal={Journal of business research},
  volume={100},
  pages={547--560},
  year={2019},
  publisher={Elsevier}
}

@article{foo2008online,
  title={Online virtual exhibitions: Concepts and design considerations},
  author={Foo, Schubert},
  journal={DESIDOC Journal of Library \& Information Technology},
  volume={28},
  number={4},
  pages={22--34},
  year={2008},
  publisher={Defence Scientific Information \& Documentation Center, Delhi}
}

@inproceedings{sakashita2024sharednerf,
  title={SharedNeRF: Leveraging photorealistic and view-dependent rendering for real-time and remote collaboration},
  author={Sakashita, Mose and Thoravi Kumaravel, Balasaravanan and Marquardt, Nicolai and Wilson, Andrew David},
  booktitle={Proceedings of the 2024 CHI Conference on Human Factors in Computing Systems},
  pages={1--14},
  year={2024}
}

@article{kerbl20233d,
  title={3d gaussian splatting for real-time radiance field rendering.},
  author={Kerbl, Bernhard and Kopanas, Georgios and Leimk{\"u}hler, Thomas and Drettakis, George},
  journal={ACM Trans. Graph.},
  volume={42},
  number={4},
  pages={139--1},
  year={2023}
}

@inproceedings{schonberger2016structure,
  title={Structure-from-motion revisited},
  author={Schonberger, Johannes L and Frahm, Jan-Michael},
  booktitle={Proceedings of the IEEE conference on computer vision and pattern recognition},
  pages={4104--4113},
  year={2016}
}

@inproceedings{chen2024mvsplat,
  title={Mvsplat: Efficient 3d gaussian splatting from sparse multi-view images},
  author={Chen, Yuedong and Xu, Haofei and Zheng, Chuanxia and Zhuang, Bohan and Pollefeys, Marc and Geiger, Andreas and Cham, Tat-Jen and Cai, Jianfei},
  booktitle={European Conference on Computer Vision},
  pages={370--386},
  year={2024},
  organization={Springer}
}

@inproceedings{brachmann2023ace,
    title={Accelerated Coordinate Encoding: Learning to Relocalize in Minutes using RGB and Poses},
    author={Brachmann, Eric and Cavallari, Tommaso and Prisacariu, Victor Adrian},
    booktitle={CVPR},
    year={2023},
}

@inproceedings{arnold2022mapfree,
	title={Map-free Visual Relocalization: Metric Pose Relative to a Single Image},
	author={Arnold, Eduardo and Wynn, Jamie and Vicente, Sara and Garcia-Hernando, Guillermo and Monszpart, {\'{A}}ron and Prisacariu, Victor Adrian and Turmukhambetov, Daniyar and Brachmann, Eric},
	booktitle={ECCV},
	year={2022},
}

@article{zhou2018stereo,
  title={Stereo magnification: Learning view synthesis using multiplane images},
  author={Zhou, Tinghui and Tucker, Richard and Flynn, John and Fyffe, Graham and Snavely, Noah},
  journal={arXiv preprint arXiv:1805.09817},
  year={2018}
}

@incollection{snavely2006photo,
  title={Photo tourism: exploring photo collections in 3D},
  author={Snavely, Noah and Seitz, Steven M and Szeliski, Richard},
  booktitle={ACM siggraph 2006 papers},
  pages={835--846},
  year={2006},
  publisher={ACM}
}

@article{wu2011visualsfm,
  title={VisualSFM: A visual structure from motion system},
  author={Wu, Changchang and others},
  journal={Proceedings of the 2011 IEEE International Conference on Computer Vision Workshops (ICCV Workshops)},
  year={2011},
  publisher={St. Louis, MO, USA}
}

@inproceedings{charatan2024pixelsplat,
  title={pixelsplat: 3d gaussian splats from image pairs for scalable generalizable 3d reconstruction},
  author={Charatan, David and Li, Sizhe Lester and Tagliasacchi, Andrea and Sitzmann, Vincent},
  booktitle={Proceedings of the IEEE/CVF conference on computer vision and pattern recognition},
  pages={19457--19467},
  year={2024}
}

@inproceedings{du2023learning,
  title={Learning to render novel views from wide-baseline stereo pairs},
  author={Du, Yilun and Smith, Cameron and Tewari, Ayush and Sitzmann, Vincent},
  booktitle={Proceedings of the IEEE/CVF Conference on Computer Vision and Pattern Recognition},
  pages={4970--4980},
  year={2023}
}

@inproceedings{suhail2022generalizable,
  title={Generalizable patch-based neural rendering},
  author={Suhail, Mohammed and Esteves, Carlos and Sigal, Leonid and Makadia, Ameesh},
  booktitle={European Conference on Computer Vision},
  pages={156--174},
  year={2022},
  organization={Springer}
}

@inproceedings{qi2017pointnet,
  title={Pointnet: Deep learning on point sets for 3d classification and segmentation},
  author={Qi, Charles R and Su, Hao and Mo, Kaichun and Guibas, Leonidas J},
  booktitle={Proceedings of the IEEE conference on computer vision and pattern recognition},
  pages={652--660},
  year={2017}
}

@inproceedings{brachmann2017dsac,
  title={Dsac-differentiable ransac for camera localization},
  author={Brachmann, Eric and Krull, Alexander and Nowozin, Sebastian and Shotton, Jamie and Michel, Frank and Gumhold, Stefan and Rother, Carsten},
  booktitle={Proceedings of the IEEE conference on computer vision and pattern recognition},
  pages={6684--6692},
  year={2017}
}

@article{brachmann2021visual,
  title={Visual camera re-localization from RGB and RGB-D images using DSAC},
  author={Brachmann, Eric and Rother, Carsten},
  journal={IEEE transactions on pattern analysis and machine intelligence},
  volume={44},
  number={9},
  pages={5847--5865},
  year={2021},
  publisher={IEEE}
}

@inproceedings{shotton2013scene,
  title={Scene coordinate regression forests for camera relocalization in RGB-D images},
  author={Shotton, Jamie and Glocker, Ben and Zach, Christopher and Izadi, Shahram and Criminisi, Antonio and Fitzgibbon, Andrew},
  booktitle={Proceedings of the IEEE conference on computer vision and pattern recognition},
  pages={2930--2937},
  year={2013}
}

@inproceedings{yu2021pixelnerf,
  title={pixelnerf: Neural radiance fields from one or few images},
  author={Yu, Alex and Ye, Vickie and Tancik, Matthew and Kanazawa, Angjoo},
  booktitle={Proceedings of the IEEE/CVF conference on computer vision and pattern recognition},
  pages={4578--4587},
  year={2021}
}

@article{geiger2024depthsplat,
  title={DepthSplat: Connecting Gaussian Splatting and Depth},
  author={Geiger, Andreas and Pollefeys, Marc and Barath, Daniel and Blum, Hermann and Wang, Fangjinhua and Peng, Songyou and Xu, Haofei},
  year={2024},
  publisher={arXiv}
}

@inproceedings{fu2024colmapfree,
  title={COLMAP-Free 3D Gaussian Splatting},
  author={Fu, Yang and Liu, Sifei and Kulkarni, Amey and Kautz, Jan and Efros, Alexei A. and Wang, Xiaolong},
  booktitle={Proceedings of the IEEE/CVF Conference on Computer Vision and Pattern Recognition (CVPR)},
  year={2024}
}

@inproceedings{ye2024noposplat,
  title={No Pose, No Problem: Surprisingly Simple 3D Gaussian Splats from Sparse Unposed Images},
  author={Ye, Botao and Liu, Sifei and Xu, Haofei and Li, Xueting and Pollefeys, Marc and Yang, Ming-Hsuan and Peng, Songyou},
  booktitle={International Conference on Learning Representations (ICLR)},
  year={2025}
}

@inproceedings{yang2024depth,
  title={Depth anything: Unleashing the power of large-scale unlabeled data},
  author={Yang, Lihe and Kang, Bingyi and Huang, Zilong and Xu, Xiaogang and Feng, Jiashi and Zhao, Hengshuang},
  booktitle={Proceedings of the IEEE/CVF Conference on Computer Vision and Pattern Recognition},
  pages={10371--10381},
  year={2024}
}

@inproceedings{tang2024lgm,
  title={Lgm: Large multi-view gaussian model for high-resolution 3d content creation},
  author={Tang, Jiaxiang and Chen, Zhaoxi and Chen, Xiaokang and Wang, Tengfei and Zeng, Gang and Liu, Ziwei},
  booktitle={European Conference on Computer Vision},
  pages={1--18},
  year={2024},
  organization={Springer}
}

@inproceedings{xu2024grm,
  title={Grm: Large gaussian reconstruction model for efficient 3d reconstruction and generation},
  author={Xu, Yinghao and Shi, Zifan and Yifan, Wang and Chen, Hansheng and Yang, Ceyuan and Peng, Sida and Shen, Yujun and Wetzstein, Gordon},
  booktitle={European Conference on Computer Vision},
  pages={1--20},
  year={2024},
  organization={Springer}
}

@inproceedings{zhang2024gs,
  title={Gs-lrm: Large reconstruction model for 3d gaussian splatting},
  author={Zhang, Kai and Bi, Sai and Tan, Hao and Xiangli, Yuanbo and Zhao, Nanxuan and Sunkavalli, Kalyan and Xu, Zexiang},
  booktitle={European Conference on Computer Vision},
  pages={1--19},
  year={2024},
  organization={Springer}
}

@article{bhat2023zoedepth,
  title={Zoedepth: Zero-shot transfer by combining relative and metric depth},
  author={Bhat, Shariq Farooq and Birkl, Reiner and Wofk, Diana and Wonka, Peter and M{\"u}ller, Matthias},
  journal={arXiv preprint arXiv:2302.12288},
  year={2023}
}

@inproceedings{ronneberger2015u,
  title={U-net: Convolutional networks for biomedical image segmentation},
  author={Ronneberger, Olaf and Fischer, Philipp and Brox, Thomas},
  booktitle={Medical image computing and computer-assisted intervention--MICCAI 2015: 18th international conference, Munich, Germany, October 5-9, 2015, proceedings, part III 18},
  pages={234--241},
  year={2015},
  organization={Springer}
}

@inproceedings{lin2017feature,
  title={Feature pyramid networks for object detection},
  author={Lin, Tsung-Yi and Doll{\'a}r, Piotr and Girshick, Ross and He, Kaiming and Hariharan, Bharath and Belongie, Serge},
  booktitle={Proceedings of the IEEE conference on computer vision and pattern recognition},
  pages={2117--2125},
  year={2017}
}

@article{gao2019pixel,
  title={Pixel transposed convolutional networks},
  author={Gao, Hongyang and Yuan, Hao and Wang, Zhengyang and Ji, Shuiwang},
  journal={IEEE transactions on pattern analysis and machine intelligence},
  volume={42},
  number={5},
  pages={1218--1227},
  year={2019},
  publisher={IEEE}
}

@article{sugawara2019checkerboard,
  title={Checkerboard artifacts free convolutional neural networks},
  author={Sugawara, Yusuke and Shiota, Sayaka and Kiya, Hitoshi},
  journal={APSIPA Transactions on Signal and Information Processing},
  volume={8},
  pages={e9},
  year={2019},
  publisher={Cambridge University Press}
}

@article{paszke2017automatic,
  title={Automatic differentiation in pytorch},
  author={Paszke, Adam and Gross, Sam and Chintala, Soumith and Chanan, Gregory and Yang, Edward and DeVito, Zachary and Lin, Zeming and Desmaison, Alban and Antiga, Luca and Lerer, Adam},
  year={2017}
}

@article{loshchilov2017decoupled,
  title={Decoupled weight decay regularization},
  author={Loshchilov, Ilya and Hutter, Frank},
  journal={arXiv preprint arXiv:1711.05101},
  year={2017}
}

@inproceedings{oufqir2020arkit,
  title={ARKit and ARCore in serve to augmented reality},
  author={Oufqir, Zainab and El Abderrahmani, Abdellatif and Satori, Khalid},
  booktitle={2020 international conference on intelligent systems and computer vision (ISCV)},
  pages={1--7},
  year={2020},
  organization={IEEE}
}

@ARTICLE{1284395,
  author={Zhou Wang and Bovik, A.C. and Sheikh, H.R. and Simoncelli, E.P.},
  journal={IEEE Transactions on Image Processing}, 
  title={Image quality assessment: from error visibility to structural similarity}, 
  year={2004},
  volume={13},
  number={4},
  pages={600-612},
  doi={10.1109/TIP.2003.819861}}

@inproceedings{zhang2018unreasonable,
  title={The unreasonable effectiveness of deep features as a perceptual metric},
  author={Zhang, Richard and Isola, Phillip and Efros, Alexei A and Shechtman, Eli and Wang, Oliver},
  booktitle={Proceedings of the IEEE conference on computer vision and pattern recognition},
  pages={586--595},
  year={2018}
}

@inproceedings{kendall2015posenet,
  title={Posenet: A convolutional network for real-time 6-dof camera relocalization},
  author={Kendall, Alex and Grimes, Matthew and Cipolla, Roberto},
  booktitle={Proceedings of the IEEE international conference on computer vision},
  pages={2938--2946},
  year={2015}
}

@inproceedings{wang2024dust3r,
  title={Dust3r: Geometric 3d vision made easy},
  author={Wang, Shuzhe and Leroy, Vincent and Cabon, Yohann and Chidlovskii, Boris and Revaud, Jerome},
  booktitle={Proceedings of the IEEE/CVF Conference on Computer Vision and Pattern Recognition},
  pages={20697--20709},
  year={2024}
}

@inproceedings{wang2025vggt,
  title={Vggt: Visual geometry grounded transformer},
  author={Wang, Jianyuan and Chen, Minghao and Karaev, Nikita and Vedaldi, Andrea and Rupprecht, Christian and Novotny, David},
  booktitle={Proceedings of the Computer Vision and Pattern Recognition Conference},
  pages={5294--5306},
  year={2025}
}
	}

\end{document}